# From Knowing to Doing: Learning Diverse Motor Skills through Instruction Learning

Linqi Ye, Jiayi Li, Yi Cheng, Xianhao Wang, Bin Liang, Yan Peng

*Abstract*— Recent years have witnessed many successful trials in the robot learning field. For contact-rich robotic tasks, it is challenging to learn coordinated motor skills by reinforcement learning. Imitation learning solves this problem by using a mimic reward to encourage the robot to track a given reference trajectory. However, imitation learning is not so efficient and may constrain the learned motion. In this paper, we propose instruction learning, which is inspired by the human learning process and is highly efficient, flexible, and versatile for robot motion learning. Instead of using a reference signal in the reward, instruction learning applies a reference signal directly as a feedforward action, and it is combined with a feedback action learned by reinforcement learning to control the robot. Besides, we propose the action bounding technique and remove the mimic reward, which is shown to be crucial for efficient and flexible learning. We compare the performance of instruction learning with imitation learning, indicating that instruction learning can greatly speed up the training process and guarantee learning the desired motion correctly. The effectiveness of instruction learning is validated through a bunch of motion learning examples for a biped robot and a quadruped robot, where skills can be learned typically within several million steps. Besides, we also conduct sim-to-real transfer and online learning experiments on a real quadruped robot. Instruction learning has shown great merits and potential, making it a promising alternative for imitation learning.

## I. INTRODUCTION

With the successful application to legged robots [1][2], aerial robots [3], and robotic manipulation [4], the learning method has been proven to be an efficient way for robot motion control. Reinforcement learning aims to maximize a reward through interactions with the environment and observations of how it responds. Although many powerful reinforcement learning methods have been proposed, such as deep deterministic policy gradient (DDPG) [5], proximal policy optimization (PPO) [6], and soft actor critic (SAC) [7], the reward design remains a challenging problem, especially for contact-rich tasks. For example, when learning legged locomotion, a simple reward may result in unnatural motion behaviors [8]. Although the robot can learn to move fast and transverse obstacles, it behaves awkwardly.

Imitation learning provides a way to generate natural-looking behaviors by applying a motion reference signal to the reward, which is usually obtained from motion capture data [9]. However, imitation learning does have some drawbacks. Firstly, the imitation process is reward-driven, where the robot starts from random actions and tries to improve it to obtain a higher reward. This process is different from the human learning process. For example, when learning to dance, we can do it the first time according to the teacher's demonstration or instruction, just not so well. Then we do it better and better with more trials, just as the saying says "practice makes perfect". In this way, we can learn sophisticated skills very efficiently. However, imitation learning usually has very bad starts and takes millions of trials to finally master the desired motion. Secondly, imitation learning tries to minimize the error between the robot's motion and the reference motion, which constrains the robot and may prevent it from exploring behaviors that have better performance. Whereas for human learning skills, we are not trying to exactly copy the behavior of others, but take it as a good start and try to make it best suit ourselves.

Inspired by the motion skill learning process of humans, we propose the instruction learning framework. The core of instruction learning is "knowing-to-doing", where the robot starts from an instructed action and improves it through training. Compared to imitation learning, which uses a reference signal in the reward, instruction learning uses the reference signal directly in the action as feedforward and applies reinforcement learning to make corrections to achieve a given goal. The learning mechanism makes instruction learning inherently more efficient and less constrained.

The contributions of this paper are fourfold. First, we propose the concept of instruction learning and elaborate on its similarities with human learning and essential differences with imitation learning. Although [10] has proposed a similar feedforward-feedback learning structure, instruction learning is different in two aspects: 1) We propose the action bounding technique by using discounted feedback, which is the key for fast learning; 2) We remove the mimic reward in instruction learning, which gives it more flexibility. By using instruction learning, the robots can learn various locomotion behaviors with a simple stepping motion as feedforward. Second, we fully demonstrate the efficiency, versatility, and flexibility of instruction learning through a bunch of locomotion learning examples, including bipedal walk, level walk, march walk, jump, and hop, as well as quadrupedal trot, pace, bound, and pronk. Third, we investigate another technique that is neglected in previous studies of imitation learning, that is,

*Research supported by the National Natural Science Foundation of China under grants No. 62003188 and No.92248304. (*Corresponding Authors: Bin Liang & Yan Peng*)

The code is available at https://github.com/Lr-2002/raisimLib.

L. Ye, Y. Peng is with the Institute of Artificial Intelligence, Collaborative Innovation Center for the Marine Artificial Intelligence, Shanghai University, 200444 Shanghai, China.

J. Li and Y. Cheng are with Center of Intelligent Control and Telescience, Tsinghua Shenzhen International Graduate School, Tsinghua University, 518055 Shenzhen, China.

X. Wang is with the School of Mechano-Electronic Engineering, Xidian University, 710126, Xi'an, China.

B. Liang is with the Navigation and Control Research Center, Department of Automation, Tsinghua University, 100084 Beijing, China.

reference in observation (RO). In most studies of imitation learning, the reference signal is put in the reward but not added to the observation, which we think is one root that causes its inefficiency. We show that RO can facilitate imitation learning to learn faster and correctly. However, even with RO, imitation learning is still not as efficient as instruction learning, which might be because of their inherent learning mechanism. Last, we demonstrate the feasibility of instruction learning on real systems by taking sim-to-real transfer and online learning experiments on a real quadruped robot. Thanks to the high efficiency of instruction learning, the robot can learn in the real world to walk forward or turn around in only 20 minutes.

## II. RELATED WORK

### A. Robot Reinforcement Learning

Reinforcement learning has become one of the mainstreams for robot locomotion control, which has led to many remarkable achievements in recent years. Duan et al. [11] propose an approach that combines bipedal robot control with reinforcement learning to allow for feet setpoint learning in task space and develop a reinforcement learning formulation for training dynamic gait controllers that can respond to specified touchdown locations. Siekmann et al. [12] index time via a cycle time variable with a periodic reward function accordingly, and common bipedal gaits including walking, running, hopping, galloping, and skipping are realized. Learning-based state estimation networks and dynamics randomization are introduced to better demonstrate fast and robust locomotion on rough terrains [13]. Margolis et al. [14] use a teacher-student training policy and a few adaptive curriculum update rules to realize fast moving in both indoor and outdoor running. Hwangbo et al. [15] leverage an attention-based recurrent encoder that integrates proprioceptive and exteroceptive input. They successfully deployed the controller on the ANYmal C robot which traversed challenging natural environments with steep inclination, slippery surfaces, grass, and snow. Margolis et al. [16] propose learning a single policy that encodes a structured family of locomotion strategies that solve training tasks in different ways evolving a conditional part to specify tasks, to avoid the slow and iterative cycle of reward and environment redesign.

Great progress based on reinforcement learning has been achieved so far, and some new frameworks are explored to accelerate the learning process. Kim et al. [17] find that leveraging constraints in the learning pipeline is promising to replace the substantial and laborious reward engineering and propose a novel framework for training neural network controllers for complex robotic systems consisting of both rewards and constraints. A reinforcement learning framework using massive parallelism on a single GPU [18] was proposed and the training time can be reduced by multiple orders of magnitude compared to previous work. It is demonstrated that a complex real-world robotics task can be trained in minutes using NVIDIA's Isaac Gym simulation environment [19].

### B. Robot Imitation Learning

For robot reinforcement learning, reward design is a crucial and challenging part. It's more intuitive for controllers like trajectory optimization methods to generate complex movements such as backflips for legged robots, while designing rewards for reinforcement learning methods to achieve this is difficult. Imitation learning is a good way to pick up pattern priori from the experts' experiences as a simple reward term, as these experiences may not be intuitive enough to form mathematical expressions to generate different motion characteristics like maintaining adequate foot clearances and behaving naturally. To some extent, imitation learning leverages expert knowledge and avoids complex reward design.

The easiest way to achieve imitation learning is to use a mimic reward to encourage the robot to follow a given motion. The mimic reward is the error between the robot state and the reference trajectory to be followed. In [9], the captured motion data from a real dog is used as reference trajectories for a quadruped robot to implement imitation learning, which successfully learns a diverse set of natural behaviors. However, imitation learning usually has low efficiency, sometimes it learns nothing. Therefore, the reference state initialization and early termination techniques are proposed in [20], which are found to be critical for achieving highly dynamic skills learning. Besides, early termination is further extended to spacetime bounds in [10], where the episodes are terminated immediately once any spacetime bounds are violated. Another problem with imitation learning is that the resulting motion is constrained to the given reference. To solve this problem, a task reward is introduced in [20] to enable adjustment of the learned behavior. In [21], a learning curriculum with imitation relaxation is designed, which can eliminate the influence of the non-ideal reference trajectory and excavate the full potential of the robot.

Behavioral cloning (BC), inverse reinforcement learning (IRL), and generative adversarial imitation learning (GAIL) also belong to imitation learning [22]. BC [23] learns a policy as a supervised learning problem and generates the state-action trajectory by matching the teaching movements. IRL [24] seeks to best explain the observed behavior by learning the corresponding reward function, and the policy is learned through an iterative loop of the Markov decision process solving and reward function updating. GAIL [25] applies the ideas of GANs [26][27] to imitate an expert. It directly extracts a policy from data and avoids significant computational expense in learning the reward function. Baram et al. [28] introduce the Model-based Generative Adversarial Imitation Learning (MGAIL) algorithm which uses relatively fewer expert samples and interactions with the environment. A forward model is given to make the computation fully differentiable, which enables training policies using the exact gradient of the discriminator. Merel et al. [29] extend GAIL to enable training for generic neural network policies and produce humanlike movement patterns from limited demonstrations. Peng et al. [30][31] propose adversarial motion priors (AMP), which combine adversarial imitation learning and unsupervised learning to develop skill embeddings that produce life-like behaviors. Task-specific annotation or segmentation of the motion data is needless and models can be trained using large datasets of unstructured motion clips. Vollenweider et al. [32] introduce multi-AMP which augments the concept of AMP to allow for multiple, discretely switchable styles. Their approach is validated in real-world experiments with a wheeled-legged quadruped robot switching between a quadrupedal and a humanoid configuration.

## C. Robot Online Learning

In a standard paradigm of robot reinforcement learning, the controller is conventionally trained first in the simulation environment and then transferred to the real robot. The sim-to-real gap which derives from the model mismatch, unpredictable environmental changes, controller delay, and other factors makes it a challenge for algorithms to work in real systems. Therefore, various techniques are proposed to solve this problem.

One solution is to increase the robustness of the policy in simulation before the algorithm deployment. Domain randomization is introduced to bridge the "reality gap" by training a policy on the distribution of environmental conditions in simulation [33]-[35]. Kumar et al. [36] present the rapid motor adaptation algorithm including an adaptation module as a subsystem to estimate the environment configuration using the information on the robot's recent state and action history. Zhuang et al. [37] distill their policy before deployment. They apply pre-processing techniques to both the raw rendered depth image and the raw real-world depth image to bridge the sim-to-real gap in visual appearance. Tan et al. [38] introduce a hierarchical framework that uses a high-level learned policy to adjust the low-level trajectory generator for better adaptability to the terrain. However, these algorithms generate relatively conservative behavior for the policy to fit both simulation and real-world environments.

Another solution is applying online learning, namely training in the real world directly. Although online learning has been successfully verified on fixed-base robots such as robotic arms through a large amount of data training [39][40], researchers emphasize the prohibitively expensive cost of tunning on some floating-base physical systems which can be easily damaged through extensive trial-and-error learning so that sample efficiency and security issue is crucial for legged robots [41]. Ha et al. [42] overcome the safety challenge by developing a safety-constrained RL framework, where the pitch and roll angles of the robot's torso are limited. Besides, robots reset after each episode is another important issue for online learning. Yang et al. [43] adopt model predictive control to account for planning latency and embed prior knowledge of leg trajectories into the action space and yield an approach that requires only 4.5 minutes of real-world data collection to learn robust and fast gaits for a quadruped robot. Bloesch et al. [44] train a small humanoid robot to walk and interact with the environment relying only on onboard sensors and limited prior knowledge of the hardware. If the robot has fallen at the end of an episode, a robust pre-programmed feedforward controller to stand up will be woken up. Gupta et al. [45] propose a reset-free method by learning multiple tasks together since different combinations of tasks can serve to perform resets for each other. This technique avoids frequent human intervention. Smith et al. [46][47] present a system that enables online locomotion policy fine-tuning via a combination of autonomous data collection and reinforcement learning.

## III. CONCEPT OF INSTRUCTION LEARNING

### A. Instruction Learning Framework

The framework of instruction learning is shown in Fig. 1. Unlike traditional reinforcement learning where the control action is entirely determined by a neural network, the control action of instruction learning consists of two parts: a feedback part learned by a neural network and a feedforward part which follows an instructed action.

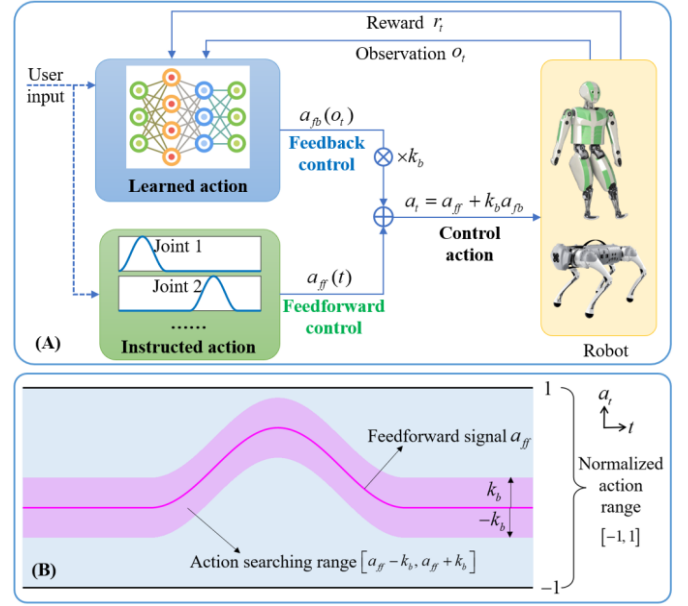

Figure 1. Instruction learning framework. (A) The feedforward-feedback control structure of instruction learning. (B) Illustration of action bounding.

The feedforward is joint angle trajectories concerning time, which is a priori knowledge. The feedback is a neural network that maps the robot's state to action, which is learned through reinforcement learning. The feedforward gives a rough routine and initial guess for the control action, while the feedback shapes it according to the reward to achieve a given goal such as maintaining balance and moving forward.

Since the control action is a combination of the feedforward and the feedback, the influence of the two components depends on their magnitudes. If the magnitude of the feedback is much larger than that of the feedforward, then the action is dominated by the feedback and thus the instruction role of the feedforward is weakened. Therefore, we introduce the action bounding technique, that is, multiplying the feedback by a ratio $k_b$ before adding it to the feedforward. The feedback ratio determines the action bounding range, which constrains the magnitude of the feedback. By adjusting the feedback ratio, the boundaries of the action are specified so we can determine how far the action can be deviated from the feedforward action, as illustrated in Fig. 1 (B). We found that the feedback ratio is crucial for the learning efficiency of instruction learning. With the action bounding technique, the searching range of the action space is greatly reduced and thus can significantly accelerate the training process.

## B. Association with Human Learning

Fig. 2 shows the instruction learning process, which is inspired by humans. One example is learning to swim, the beginner has a prior knowledge of how to move the limbs in sequence but does not know how to coordinate the movements according to the body state in the water. The prior knowledge can be seen as feedforward and the coordination is feedback that needs to be learned from interaction with the water. Another example is the babies learning to walk. It is known that babies can take steps when held upright with the feet touching a solid surface, which is called the "stepping reflex" [48]. The stepping reflex is an important part of babies' development, which prepares them to take real steps several months down the road. The stepping reflex can also be viewed as feedforward, which is prior knowledge inherited from parents and works along with postnatal training to help babies walk.

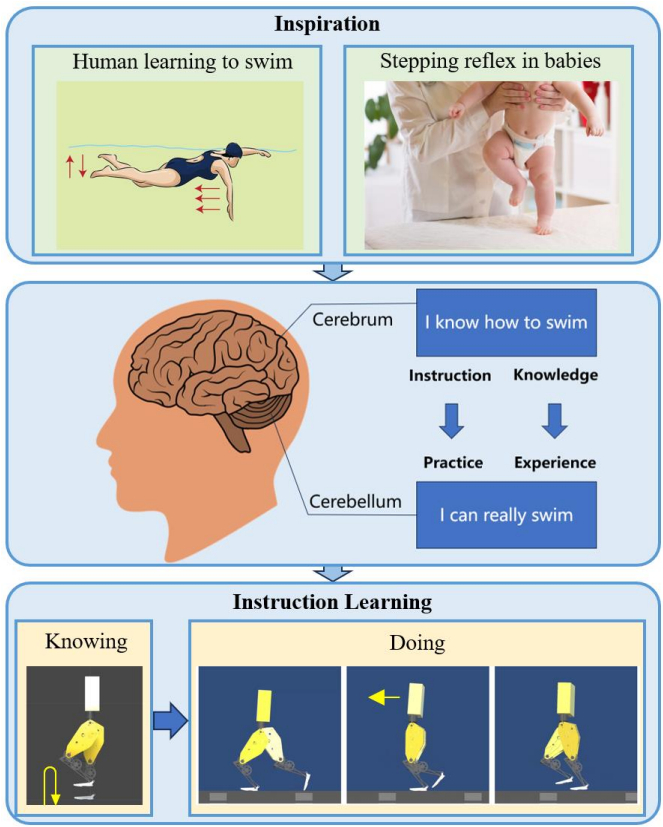

Figure 2. Association of instruction learning with human learning.

The counterparts of the feedforward and the feedback are knowledge and experience in the human brain, which are related to the cerebrum and cerebellum, respectively. For various human motion skills like swimming, dancing, bicycling, skiing, etc., we usually know of it first (usually from instruction), and then become experienced in it through practice. Similarly, the process of instruction learning is also from knowing to doing. For example, for a biped robot learning to walk, we can use the stepping motion as a feedforward. However, that is not enough for walking, the robot will fall if we put it on the ground. A feedback action is needed to stabilize the motion, and this is exactly what the reinforcement learning part is doing. Through massive trials and interactions, the robot learns to maintain balance, it can step in place, and can also walk at a given speed. In this learning process, the feedforward is very important for the robot to converge to a desired walking style.

Compared to the other reinforcement learning methods, instruction learning introduces an additional feedforward in the controller. Therefore, we now have two degrees of freedom during control design, that is, the reward and the feedforward. The feedforward directly affects the controller, while the reward gradually shapes the controller through training. Tunning the feedforward and reward helps us achieve different goals. For example, the feedforward can achieve periodic motion, a predefined motion sequence, or a given pose easily, while the reward can help maintain balance and motion coordination.

## C. Comparison with Imitation Learning

Instruction learning and imitation learning both aim at reproducing some given motion styles, but their mechanisms are indeed very different. To illustrate this, we made a vivid comparison in Fig. 3.

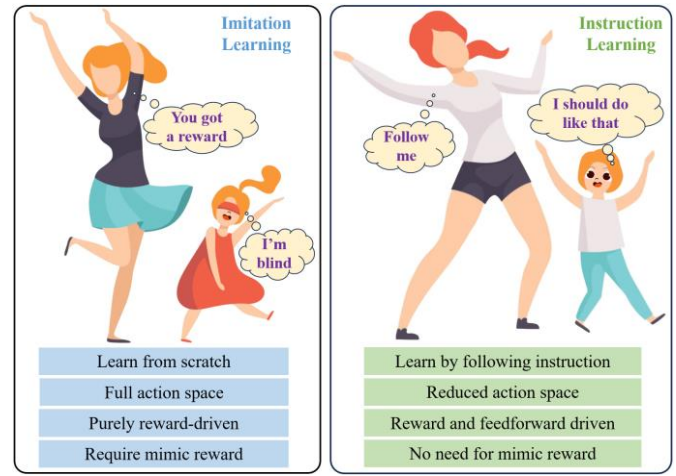

Figure 3. Imitation learning vs. instruction learning.

Imitation learning is like blind people learning to dance, the learner does not know the dancing movements in mind (the observation). It learns from scratch, explores in a full action space and the action is purely reward-driven. For instruction learning, the learner builds the concept of dancing by watching the teacher's movement, and then it learns by following instructions/demonstrations. It explores in a reduced action space around the instructed/demonstrated action. Its action is driven by both reward and feedforward. Therefore, the learning mechanism natively determines that instruction learning can be more efficient than imitation learning.

## IV. INSTRUCTION LEARNING FOR LEGGED LOCOMOTION

To verify the effectiveness of instruction learning, we take a biped robot and a quadruped robot as examples and apply instruction learning to learn different locomotion behaviors.

### A. Robot Model

The simulated robot models are shown in Fig. 4. For the biped robot, we use a modified model of Ranger Max (previously called Cornell Tik-Tok [49]). The real Ranger Max only has four joints in each leg. To enhance its capability, we add the hip yaw and ankle roll joints to the simulation model. For the quadruped robot, we use the Unitree A1 model. Specifications for the two robots are as follows.

Biped robot: Each leg has 6 actuated joints. The range of motion (take the position in Fig. 4 as zero) for each joint is (unit: degree): hip yaw -30~30, hip roll -30~30, hip pitch -60~120, knee pitch -170~10, ankle roll -30~30, ankle pitch -60~60. The maximum joint force is 200 Nm. The joint position control loop has a stiffness of 2000 and a damping of 100.

Quadruped robot: Each leg has 3 actuated joints. The range of motion (take the position in Fig. 4 as zero) for each joint is (unit: degree): hip roll -46~46, hip pitch -90~210, knee pitch -94.5~7.5. The maximum joint force is 33.5 Nm. The joint position control loop has a stiffness of 180 and a damping of 8.

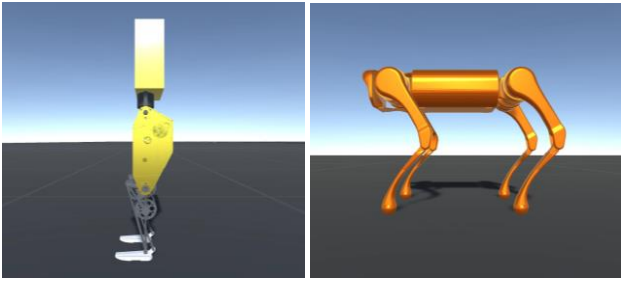

Figure 4. The simulated robot models. Left: Ranger Max; right: Unitree A1.

### B. Feedforward Design

The feedforward is the position reference trajectory for each joint. For simplicity, we consider two commonly used trajectories.

The first is a periodic trajectory, where the joint moves cyclically with a fixed period. In this case, the reference angle is designed as a sinusoidal signal

$$\theta_t^{ref} = \theta_0 + \Delta\theta \frac{1-\cos(2\pi t/T)}{2} \quad (1)$$

The second is a point-to-point trajectory, where the joint moves from one position to another position. In this case, the reference angle is designed as a ramp signal

$$\theta_t^{ref} = \theta_0 + \Delta\theta \frac{t}{T} \quad (2)$$

The feedforward action is obtained by mapping the reference angle to -1~1:

$$a_{ff} = 2\frac{\theta_t^{ref} - \theta_{\min}}{\theta_{\max} - \theta_{\min}} - 1 \quad (3)$$

where $\theta_t^{ref}$ is the reference angle of the joint, and $\theta_{\min}, \theta_{\max}$ represent the lower limit and upper limit of the joint range of motion, respectively.

For legged locomotion, an important feature is the foot swings up and down periodically. Therefore, we decided to use the foot swing motion as feedforward. For legged robots, by using different foot swing sequences and swing trajectories, various gaits can be easily learned. In this paper, we explore various gaits including bipedal walking, level walking, march walking, hopping, and jumping, as well as quadrupedal trotting, pacing, bounding, and pronking. The detailed feedforward action for each gait can be found in the attached video. For most gaits, we only use the sinusoidal trajectory (1) to generate the feedforward. Only for biped march walking and biped hopping, we have applied a combination of the sinusoidal trajectory (1) and the ramp trajectory (2) to design the feedforward.

### C. Learning Setup

- Hyperparameter

We use PPO for reinforcement learning. The hyperparameters are: batch_size 2048, buffer_size 20480, learning_rate 0.0003, beta 0.005, epsilon 0.2, lambda 0.95, num_epoch 3.

- Neural network

The neural network has an actor-critic structure, where the actor network is a multi-layer perceptron (MLP) with 3 hidden layers with 512 hidden units for each layer and the critic network is an MLP with 2 hidden layers with 128 hidden units for each layer. The output of the actor network is clipped to -1~1.

- Observation

The observations are

$$o_t = [q_t \ \omega_t \ v_t \ \theta_t \ \dot{\theta}_t]^T \quad (4)$$

When RO is applied, the observations are

$$o_t = [q_t \ \omega_t \ v_t \ \theta_t \ \dot{\theta}_t \ \theta_t^{ref}]^T \quad (5)$$

where $q_t$ is the orientation of the body expressed by a quaternion, containing 4 variables. $\omega_t, v_t$ are the body angular velocity and body linear velocity in the body frame, respectively, each containing 3 variables. $\theta_t, \dot{\theta}_t, \theta_t^{ref}$ are the joint angle, joint angular velocity, and joint reference angle, respectively, each containing 12 variables (for both biped and quadruped robots).

- Action

To enhance the smoothness of the action, the output of the actor network is passed through a low-pass filter to obtain the feedback action. Denote the original output of the actor network as $a_{nn}$, then the feedback action is obtained by

$$a_{fb} = 0.9 a_{fb}^{last} + 0.1 a_{nn} \quad (6)$$

where $a_{fb}^{last}$ is the value of $a_{fb}$ in the last time step. Then the feedback action $a_{fb}$ is weighted by $k_b$ and added with the feedforward to obtain the control action

$$a_t = a_{ff} + k_b a_{fb} \quad (7)$$

The control action $a_t$ should be within the range -1~1. Since $a_{ff}$, $a_{fb}$ are both within -1~1, we choose $k_b$ within 0~2. However, the obtained $a_t$ may be out of the range -1~1. Therefore, we clip it to calculate the command joint angle

$$\theta_t^{cmd} = \theta_{\min} + \frac{\text{sat}(a_t)+1}{2}(\theta_{\max} - \theta_{\min}) \tag{8}$$

where sat() represents the linear saturation function. The command angle $\theta_t^{cmd}$ is sent to the position control loop of each joint.

*D. Reward Design*

Due to the application of the feedforward, there is no need for a tedious reward design. Indeed, we found a simple reward is enough to generate various natural-looking gaits with instruction learning. We design the reward as a combination of several components. Each component is explained as follows

1) The mimic reward $r_m$ is designed as $r_m = 1.2 - 0.02\sum_{1}^{12}\left|\left(\theta_i - \theta_i^{ref}\right)\right|$ (angle unit: degree), which encourages the joint to track the reference trajectory.

2) The alive reward $r_a$ is designed as $r_a = 1$. The robot receives this reward per time step for not falling.

3) The balance reward $r_b$ is designed as $r_b = -0.1\left(\left|\theta_{pitch}\right| + \left|\theta_{yaw}\right| + \left|\theta_{roll}\right|\right)$ (angle unit: degree), which encourages keeping the body upright.

4) For the velocity reward, we give three options

- $r_v^{step} = -\left|v_{forward}\right| - \left|v_{lateral}\right| - \left|v_{vertical}\right|$
- $r_v^{walk} = v_{forward} - \left|v_{lateral}\right| - \left|v_{vertical}\right|$
- $r_v^{jump} = v_{forward} - \left|v_{lateral}\right| + \left|v_{vertical}\right|$

where the first encourages the robot to keep its body still; the second encourages moving forward; and the third encourages moving forward and meanwhile moving up and down.

5) The synchronization reward $r_s$ is designed as $r_s = -0.05\sum_{1}^{3}\left|\left(\theta_i^{left} - \theta_i^{right}\right)\right|$ (angle unit: degree), which encourages the biped robot to synchronize the motion of the two legs ( $\theta_i$ represents pitch angle of hip, knee, and ankle).

A combination of the rewards above can be used to generate different gaits. Table I lists the rewards we applied for some commonly seen gaits.

**Table 1.** Reward design.

| Gaits | Reward |
|---|---|
| Stepping (quadruped and biped) | $r = r_m + r_a + r_b + r_v^{step}$ |
| Biped: walk, level walk, march walk<br>Quadruped: trot, pace, bound | $r = r_a + r_b + r_v^{walk}$ |
| Biped: hop<br>Quadruped: pronk | $r = r_a + r_b + r_v^{jump}$ |
| Biped: jump | $r = r_a + r_b + r_v^{jump} + r_s$ |

**Remark:** It can be seen from Table I that only the stepping motion has adopted a mimic reward. Instruction learning actually does not require a mimic reward to learn a motor skill. For the stepping motion, we add the mimic reward just to make a fair comparison with imitation learning. Removing the mimic reward will not affect instruction learning to learn the stepping motion.

Termination condition: we end the episode if the robot "falls" ( $\left|\theta_{pitch}\right| > 10\deg$ or $\left|\theta_{roll}\right| > 10\deg$ ) or if the time step reaches 1000.

V. SIMULATION RESULTS

We use Unity ML-Agents for simulation, which adopts the Physix engine. The time step for each action is 0.01 seconds, which indicates a control frequency of 100 Hz. The training is running on the CPU. To speed up the training, we use 12 copies of the agents for parallel training. In our experience, it generally takes about 10 seconds to complete 10k training steps.

*A. Stepping Motion Learning*

We first test the stepping motion learning. The aim is to let the robot learn to step in place while maintaining balance. The feedforward is a periodic stepping motion (each foot goes up and down when the robot is hanging up but it will fall when put on the ground).

● **Instruction Learning vs. Imitation Learning**

We compare four learning methods: imitation learning (IML), imitation learning with reference in the observation (IML-RO), instruction learning (INL), and instruction learning with reference in the observation (INL-RO). For IML and IML-RO, the action is $a_t = a_{fb}$. For INL and INL-RO, the action is $a_t = 0.5a_{fb} + a_{ff}$. To make a fair comparison, we use the same reward (with a mimic reward, see Table I) for all of them. The reward curves are shown in Fig. 5 for the biped robot and Fig. 6 for the quadruped robot.

From the two figures, we have the following observations:

1) For instruction learning, the influence of RO can be neglected as can be seen that the learning curves of INL and INL-RO are almost the same. This might be because the reference already enters the controller through the feedforward, therefore it is no need to add the reference in the observation.

2) RO is critical for imitation learning. Without RO, imitation learning can hardly learn the stepping motion as can be seen that the reward of IML converges to a much smaller value (the robot actually learns to stand still).

3) Instruction learning is much more efficient than imitation learning, even when RO is applied. Comparing the results of INL with IML-RO, it can be seen that during the initial learning process, INL spends much fewer time steps to achieve the same reward, especially for the quadruped robot. This indicates the high efficiency of instruction learning. Moreover, INL still maintains a higher reward even at the post-learning stage.

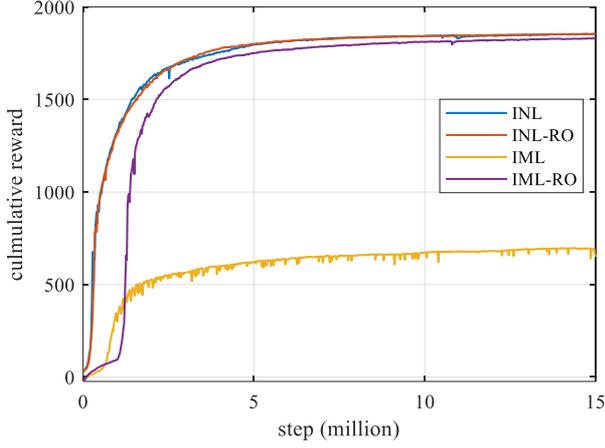

Figure 5. Reward curves for biped stepping.

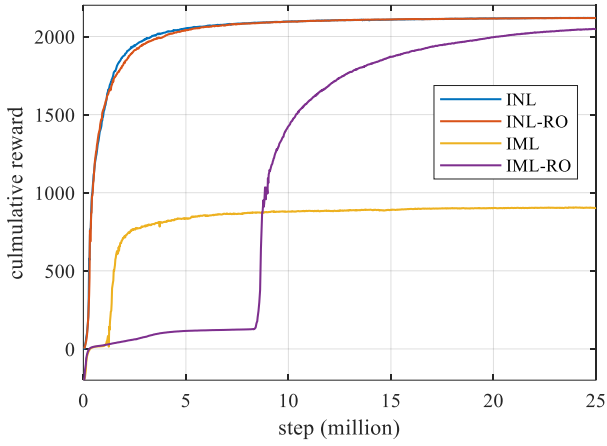

Figure 6. Reward curves for quadruped stepping.

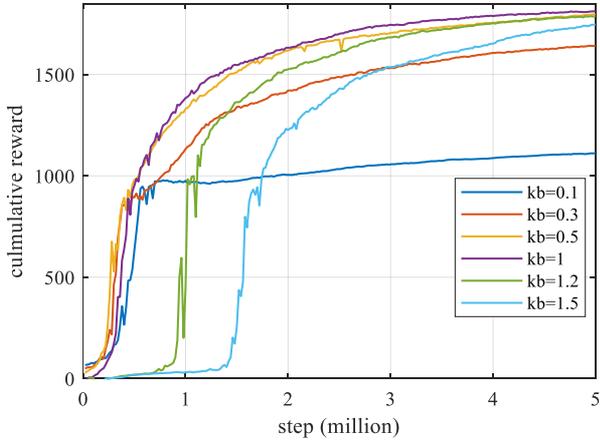

Figure 7. Reward curves for biped stepping with different $k_b$.

● **Influence of the action bounding range**

To investigate the influence of the action bounding range, we take the biped robot as an example and select different $k_b$ values. The learning results are shown in Fig. 7. It can be seen that the action bounding range has a significant influence on the learning performance. If the range is too big (see $k_b = 1.2$ and $k_b = 1.5$), then the guiding role of the feedforward reduces a lot and the learning notably slows down. If the range is too small (see $k_b = 0.1$ and $k_b = 0.3$), the learning is fast in the beginning but converges to a lower reward. This phenomenon is because the action range is too small to ensure good reference tracking, which ends up in a small mimic reward. When the range is properly selected (see $k_b = 0.5$ and $k_b = 1$), the learning is fast and the reward converges to a high value.

*B. Locomotion Learning*

In the previous stepping motion learning, we adopted a mimic reward. The function of the mimic reward is to encourage the robot to follow a reference trajectory. For imitation learning, the mimic reward is indispensable and crucial to learning the reference motion correctly. However, for instruction learning, the reference motion is directly used in the feedforward action, so there is no need for the mimic reward. In fact, there are many benefits of removing the mimic reward. Firstly, it is difficult to track a trajectory accurately, which is time-consuming for training and decreases learning efficiency. Secondly, the mimic reward constrains the motion. Removal of the mimic reward can better release the robot's potential to achieve other goals rather than just following the reference.

Therefore, for the next locomotion learning tasks, we no longer use the mimic reward or RO. With the same feedforward as the stepping motion, locomotion can be easily achieved by simply altering the velocity reward (see Table I for details). As a result, the biped robot learns to walk and the quadruped robot learns to trot. Moreover, by reorganizing the feedforward signal, the robots can easily learn other gaits. For the biped robot, we demonstrate learning of walking, level walking, march walking, jumping, and hopping. And for the quadruped robot, we demonstrate learning of trotting, pacing, bounding, and pronking. Fig. 8 and Fig. 9 show the reward curves, where we only train for 1 million steps for the biped robot and 3 million steps for the quadruped robot, which are enough to learn all the desired gaits. This indicates that instruction learning is highly efficient. Snapshots of the resulting gaits are shown in Fig. 10 and Fig. 11, respectively.

The results also verify the versatility of instruction learning. By using different feedforward signals, the robots can learn diverse locomotion tasks with simple (and even the same) rewards. Besides, compared to imitation learning, the reference signal in instruction learning can be much simpler, it only roughly characterizes the motion style and does not need to be very exact, therefore it can be manually designed instead of using motion capture data.

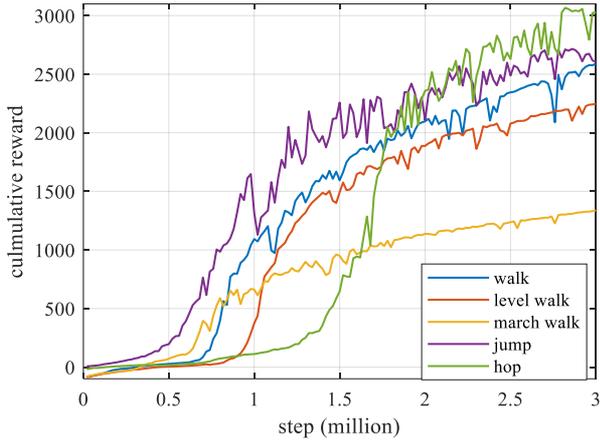

Figure 8. Reward curves for biped locomotion.

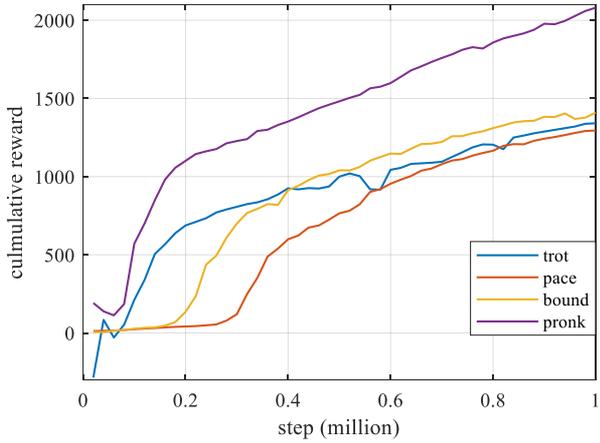

Figure 9. Reward curves for quadruped locomotion.

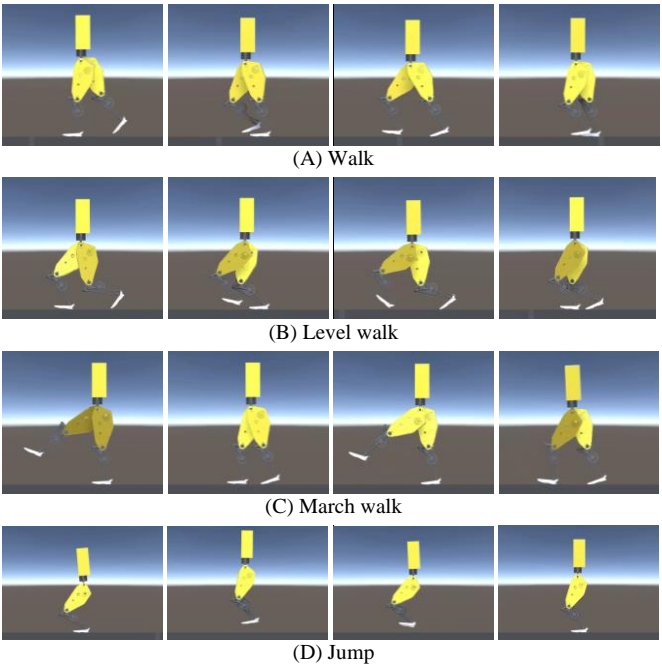

(A) Walk

(B) Level walk

(C) March walk

(D) Jump

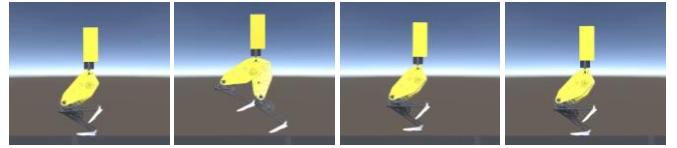

(E) Hop

Figure 10. Biped learned locomotion behavior.

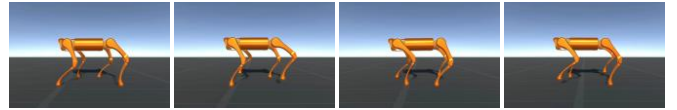

(A) Trot

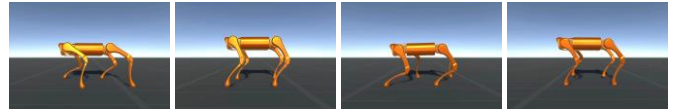

(B) Pace

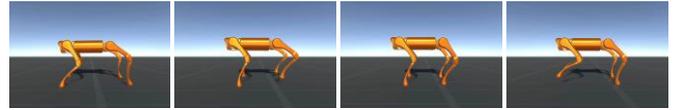

(C) Bound

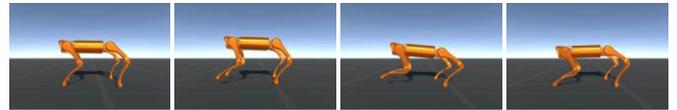

(D) Pronk

Figure 11. Quadruped learned locomotion behavior.

*C. Motion Adaptation*

Despite the superior learning efficiency, another advantage of instruction learning is its flexibility. For traditional reinforcement learning, since the neural network weights cannot be manually tuned, if we want to adjust the learned motion, the only way is to retrain the neural network. But for instruction learning, there is a feedforward that can be directly adjusted after training. Therefore, we can adjust the learned motion conveniently. As we tested, the learned motion can be maintained when the feedforward is modified in a certain range.

To illustrate this, we use the period and amplitude of the feedforward (corresponds to step period and step height for walking) as examples. When training is completed for a gait, we change the period or amplitude (independently) of the feedforward bit by bit until the gait cannot be maintained (the robot falls or does not move forward), and then we record the adaption range, which is shown in Fig. 12. It can be seen that for all gaits, a feasible range exists. The adaption range can even be as low as 0.2 and as high as 4. For some gaits like biped march walking, jumping, and hopping, the adaption range is smaller compared to other gaits, this might be because those gaits are more unstable.

The motion adaption property of instruction learning is very useful, allowing us to directly adjust the learned motion and thus eliminate the retraining time. The adaption capability is also similar to human skill learning. Once we learn a motor skill, we can easily master the variations of it. For example, we can change the gait frequency and gait clearance height easily without relearning.

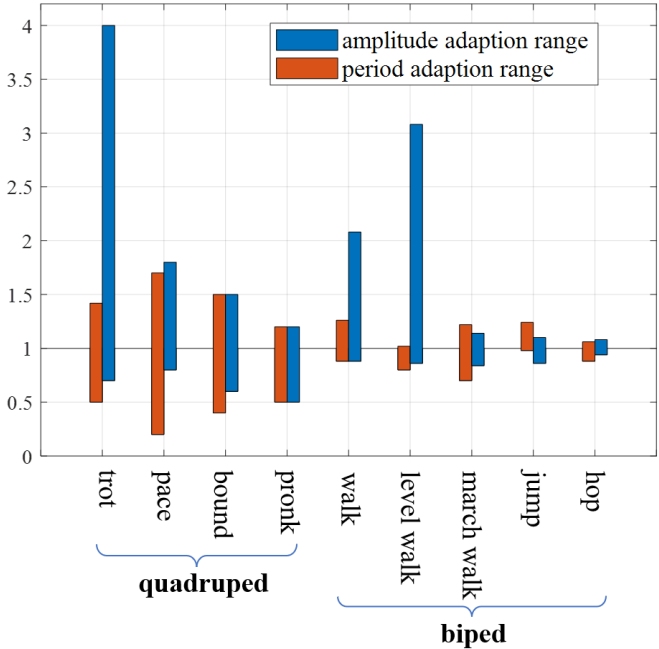

Figure 12. Adaption range of learned motion to a modified feedforward.

## VI. EXPERIMENTAL RESULTS

To verify the effectiveness of instruction learning for real systems, we do sim-to-real transfer and online learning experiments on real quadruped robots. Compared to the previous simulations, a slight modification is made to the observation. First, we remove body velocity from the observation since it is not so accurate. Second, the body orientation quaternion in the observation is replaced by the body pitch and roll angles to ensure consistency when the robot faces different directions.

### A. Sim-to-Real Transfer

We have two groups working on the sim-to-real transfer. One group uses the Raisim software and the Unitree A1 robot, while the other group uses the IsaacGym software and the Unitree Go1 robot.

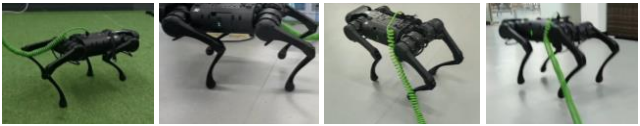

Figure 13. A1 robot at different gaits. Left to right: trot, pace, bound, pronk.

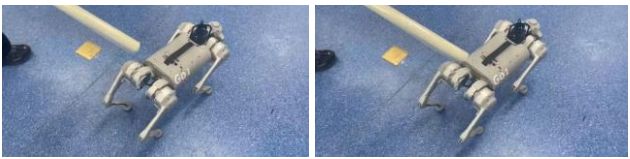

Figure 14. GO1 robot robust walking.

The first group focuses on realizing the four categories of gaits. We first try to directly transfer the learned policy to the real robot but the robot cannot keep balance due to the sim-to-real gap. We then add random disturbances to the robot during training, which results in a successful transfer, but the motion is not so robust (see Fig. 13 and the video). The second group focuses on command following and robust walking. We add the body forward velocity command and yaw velocity command in the observation and the reward and apply domain randomization during training. As a result, the robot can follow a given velocity command and maintain balance when disturbed (see Fig. 14 and the video). It can also adapt to the feedforward modification in a certain range.

### B. Online Learning

Due to the high learning efficiency, instruction learning is very well suited for real robot online learning. Online learning collects data from the real robot and updates the control policy in real time, which does not have the sim-to-real problem. However, compared to learning in simulation, online learning cannot be accelerated or trained parallelly, therefore it particularly requires highly efficient learning. Besides, resetting is troublesome in online learning when the robot falls. Instruction learning can handle these two problems thanks to its high efficiency and reduced action space which can prevent the robot from falling.

Recall the previous simulation results in Fig. 9 where the quadruped robots can learn various gaits in one million steps (time step 0.01s), which equals 2.78 hours in real time. That is about the battery life on one charge and is acceptable. However, the robot will easily fall with the feedback ratio of 0.5. Besides, as the robot walks faster and faster, we need a big enough field for the experiment. Therefore, we decided to use a smaller feedback ratio and train for a shorter time.

We select the trotting gait as an example. The robot starts from a stepping-in-place motion and learns to turn or walk forward. Before each experiment, we try a more realistic simulation first to ensure the feasibility. The reset is removed in the simulation, and the reward only uses available data the same as the real robot. When the simulation is passed, we then apply online learning to the real robot. The experimental scene is shown in Fig. 15.

During training, the robot learns on intermittent episodes. Each episode takes about 5.5 s and includes three stages: 1) The robot walks for about 3.3 s; 2) PPO updates; 3) The robot takes several steps to recover to the initial state. Both online learning tasks take about 20 minutes for training. The robot does not fall during the training and we only need to reposition it when necessary. The reward curves for the two online learning experiments are shown in Fig. 16.

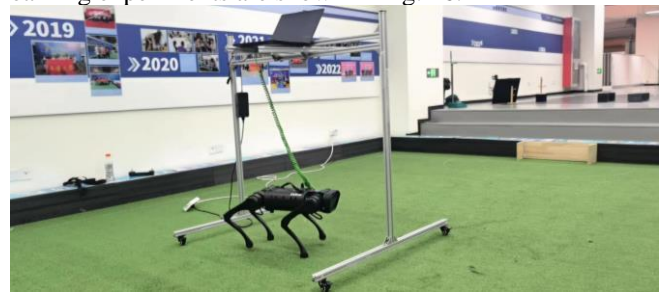

Figure 15. Online learning scene.

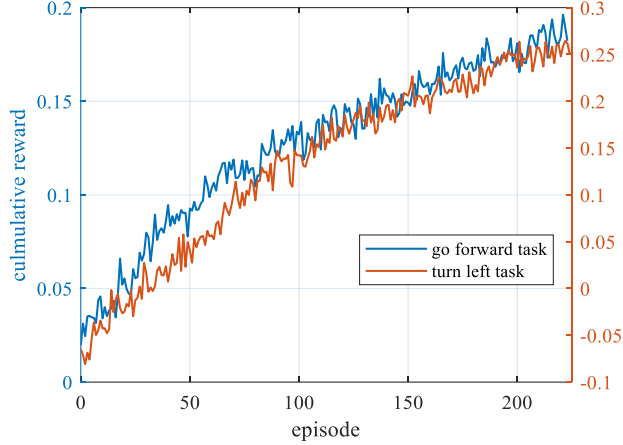

Figure 16. Reward curves for online learning.

For the turning task, the yaw velocity of the robot which can be obtained from the IMU is used as the reward. To guarantee fast learning, we apply a feedback ratio of 0.15 for the hip roll joints and 0 for the other joints. As can be seen from the video, the robot has a turning right trend in the beginning but successfully learns to turn left in the end.

For the walking forward task, the feedback ratio is 0 for the hip roll joints and 0.07 for the other joints. A problem in this task is that the reward requires to know the linear velocity of the robot but it cannot be obtained directly. We have tried to use a Kalman filter to estimate the velocity, which fuses the acceleration reading, the encoder reading, and the contact information, but the estimation performance is bad (especially when the velocity is small), which degrades the learning performance. Therefore, we give up using the velocity in the reward but instead use the angular velocity of the leg. We denote the leg angle as $\theta_{leg} = \theta_{hfe} + \theta_{kfe}/2$ ( $\theta_{hfe}, \theta_{kfe}$ represent the hip pitch and knee pitch angle, respectively), which is the pitch angle between the robot body and the virtual leg connecting the foot and the hip. Then the leg angular velocity is calculated as the leg angle difference between two time steps $\omega_{leg} = (\theta_{leg})_{k+1} - (\theta_{leg})_k$. Then the reward is designed as $r = \omega_{sp_1} + \omega_{sp_2} - \omega_{sw_1} - \omega_{sw_2}$, where $\omega_{sp_1}, \omega_{sp_2}$ represent angular velocity of the two supporting legs and $\omega_{sw_1}, \omega_{sw_2}$ the two swing legs, respectively. Using this reward, the robot successfully learns to walk forward in 20 minutes.

## VII. CONCLUSION

This paper proposes a novel reinforcement learning framework named instruction learning for robot motor skill learning and verifies its effectiveness through a bunch of locomotion learning examples. The key components of instruction learning are the feedforward-feedback learning structure, the action bounding technique, and the mimic reward elimination idea, which distinguish it from imitation learning and are critical for achieving highly efficient and flexible learning. The feedforward and feedback form a knowledge and data dual-driven learning system. The feedforward is knowledge-driven, which can be modified immediately, and the feedback is data-driven, which is learned gradually. The training process of instruction learning is like human practicing, which is a process from knowing to doing. The action bounding technique constrains the action searching range in a reduced space along the feedforward action and makes instruction learning highly efficient. Besides, instruction learning uses a simple reward and removes the mimic reward. It tries to learn a motor skill by following instructions rather than following the reference trajectory accurately. This gives instruction learning high flexibility. Finally, instruction learning also shows good adaptivity, where the learned behavior can adapt to a group of modified feedforward.

Instruction learning has many similarities to human learning, which provides an interesting perspective for us to view the motor skill learning mechanism of humans. Instruction learning is easy to understand and implement, which provides a valuable framework for efficient motion learning and has great potential for robot learning in the real world. It should be noted that we did not pay special attention to improving the robustness of the controller, so the behaviors learned by instruction learning are currently not as stable as state-of-the-art learning controllers. However, instruction learning can be combined with other sim-to-real methods to enhance the robustness of the learned controllers, which is necessary for more complex real-world applications.

ACKNOWLEDGMENT

Thanks for the insightful discussion with Professor Andy Ruina. Thanks for the experimental facilities provided by the National Demonstration Center for Experimental Engineering Training Education, Shanghai University.